# Design, Mechanical Simulation and Implementation of a New Six-Legged Robot

Ahmad Ghanbari[*#], Mahdieh Babaiasl[#], Ako Veisinejad[#]
[*]Faculty of Mechanical Engineering, University of Tabriz, 22Bahman Blvd, 5166616471 Tabriz
[#] School of Engineering Emerging Technologies, Mechatronic Research Lab, 22Bahman Blvd, 5166616471 Tabriz
Corresponding Author:
Mahdieh Babaiasl, E-mail: m.babaiasl@gmail.com
Tel: +982177122857
Fax: +984113393781

**Abstract.**

Ants are six-legged insects that can carry loads ten times heavier than their body weight. Since having six-legs, they are intrinsically stable. They are powerful and can carry heavy loads. For these reasons, in this paper a new parallel kinematic structure is proposed for a six-legged ant robot. The mechanical structure is designed and optimized in Solidworks. The mechanism has six legs and only two DC motors actuate the six legs so from mechanical point of view the design is an optimal one. The robot is light weight and semiautonomous due to using wireless modules. This feature makes this robot to be suitable to be used in social robotics and rescue robotics applications. The transmitter program is implemented in supervisor computer using LabVIEW and a microcontroller is used as the main controller. The electronic board is designed and tested in Proteus Professional and the PCB board is implemented in Altium Designer. Microcontroller programming is done in Code Vision.

**Keywords:** Ant locomotion, Six-legged robots, Hexapods, parallel robots

## 1. Introduction

Autonomous robots can be divided into manipulators and mobile robots. Based on locomotion on the ground, mobile robots can be further divided into wheeled robots and legged robots. Legged robots are characterized by the number of their legs. Among legged robots, six-legged robots have an important place since they are compliant and stable. This class of legged robots draws special attention from the researchers all over the world, because of its fascinating nature. SILO6 [1] with the application of removing landmines, underactuated hexapedal robot [2] inspired by the locomotion of a cockroach, The Space Climber [3], a bio-inspired six-legged robot [4] inspired by the walking insects, LAURON with a novel leg mechanism [5] with four degrees of freedom (DOFs), RiSE [6], AQUA [7], DASH [8], DLR-Crawler [9-10], Screenbot [11], Little Crabster [12], RHex [13], a compliant hexapod robot [14], Biologically Inspired Legged Locomotion Ant Prototype (BILL-Ant-p) [15] are some of legged robots. A complete survey of legged robots is given in [16]. Most six-legged robots are inspired by insects whose legs have four degrees of freedom. However most six-legged robots' legs possess three or less degrees of freedom that reduces their mobility. Using four DOFs for each leg enlarges the workspace of the leg and makes it redundant. Ant is one of the creatures that can carry loads ten times heavier than its body weight. This paper demonstrates design, mechanical simulation and implementation of a new six-legged robot called AntBot I inspired by ants. Initially mechanical design and simulation of the robot using Solidworks is presented and mechanical parameters of the robot are extracted. The new proposed kinematic structure has parallel mechanism. The actuators are chosen based on the maximum torques required. The two actuators drive the six legs, so from mechanical point of view the design is optimal. Afterwards mechanical implementation and electronic implementation are presented. The electrical circuit and Printed Circuit Board (PCB) of the robot are designed and tested using Proteus Professional and Altium Designer respectively. To make the robot autonomous, two wireless modules are used and the transmitter's program is programmed and tested using LabVIEW software. Atmega16 is used as the main controller and its programming is done using Codevision [17]. Finally the overall implemented robot that can walk and turn semiautonomously is shown.

## 2. Mechanical Design and simulation

### 2.1 Mechanical design using Solidworks

Figure1 (left) depicts the ultimate design of the robot in Solidworks. Figure 1 (right) through Figure 3 depict one cycle of robot locomotion. Legs are shown with different colors for better visualization. To mount IR sensor or camera, a head is also designed for the robot. The head can also be used to mount a manipulator.

### 2.2 Choosing chassis and legs material

The chassis and legs of the robot can be made from different materials. Wood, plastic, fiberglass and aluminum are good candidates for chassis and legs material. After studying the properties of these materials, the chassis and legs of the robot are chosen to be aluminum due to its light weight, low cost and corrosion resistance. To provide friction under legs of the robot and to make the locomotion of the robot smoother and easier and also to smoothen under legs and obstacle avoidance, interfacial adhesive is used.

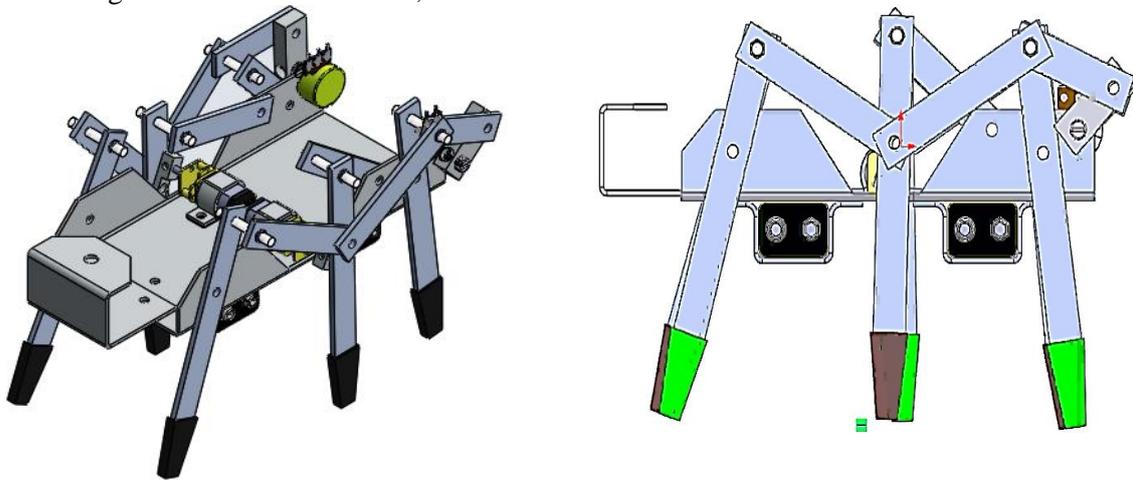

Figure 1 Ultimate mechanical design in Solidworks (left) and First stage of robot locomotion (right)

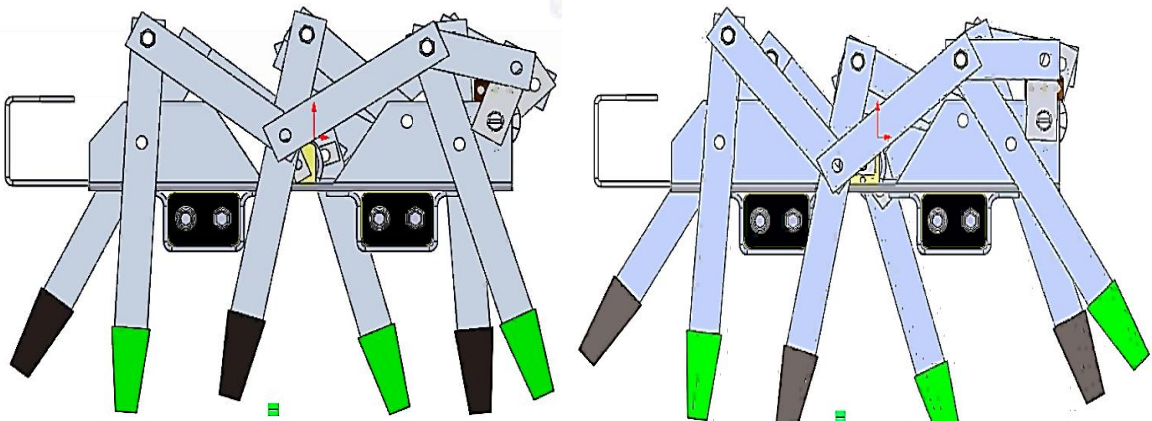

Figure 2 Second stage of robot locomotion (left) and Third stage of robot locomotion (right)

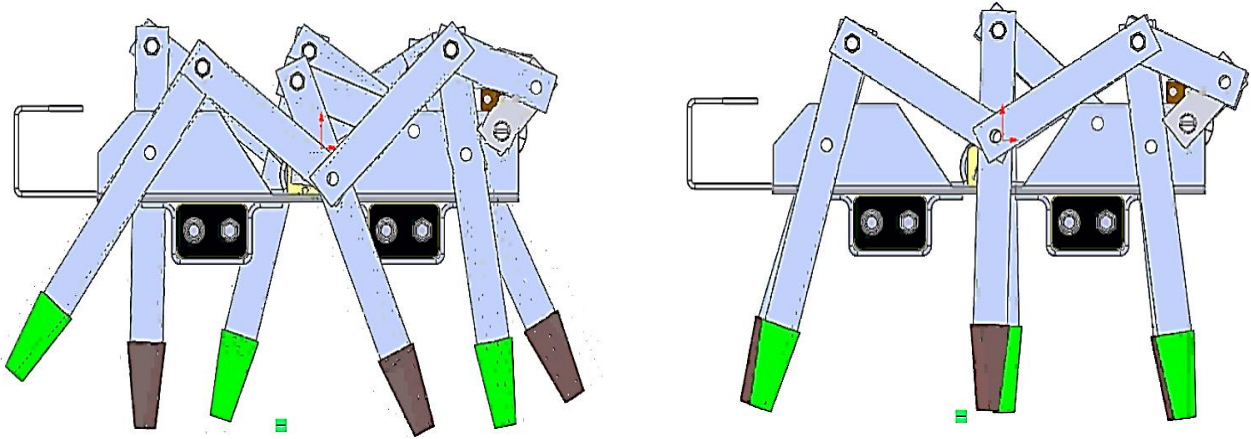

Figure 3 Fourth stage of robot locomotion (left) and Fifth stage of robot locomotion (right)

### 2.3 Choosing actuators

This robot has a minimal number of actuators with only two actuators driving the six legs. Its unique mechanism allows one actuator to drive the first three legs and one actuator to drive the other three legs. The actuators are selected based on the maximum torque. The used actuators are two 6 to 12 volts RA geared DC precious metal-brush motors.

### 2.4 Choosing power supply

The power supply is chosen based on the used motors. Due to the semiautonomous nature of the robot, the power supply should be portable. Because of this reason, two MOTOMA 9 volt batteries are chosen as power supplies.

### 2.5 Choosing sensors

Position feedback to controllers can be achieved by using two $5k\Omega$ potentiometers. For our application potentiometer can be a cost effective choice.

### 2.6 Center of mass of the robot

One of the important issues to be considered in making a robot is its stability. As shown in Figure 4, the center of mass of the robot is located on the center of the line connecting the shaft centers of two motors. So from mechanical viewpoint, the robot is stable. The overall mass of the robot without electrical board is $230 grams$. Center of mass of the robot with respect to output coordinate frame is $(-5.73, -9.25, 4.13) mm$. And the principal moments of inertia (moments of inertia taken at the center of mass and with respect to the coordinate frame attached to the center of mass) [18] are $I_{xx} = 0.000419 (kgr.m^2)$, $I_{yy} = 0.000768 (kgr.m^2)$ and $I_{zz} = 0.000931 (kgr.m^2)$.

### 3. Implementation of the robot

After CAD modeling and initial tests in the CAD environment, the final prototype of the robot is implemented in the Mechatronics research laboratory at the School of Engineering Emerging Technologies, University of Tabriz. The electronic circuit is designed and tested using Proteus Professional. Figure 5 shows the designed circuit in Proteus Professional. In the schematic, J3 is 12(v) supply input and using two regulators, the required 5(v) input to logical circuits and 6(v) input to motors are provided. J7 is power supply on/off switch. J6 is the receiver's input to the board. J4 and J5 are potentiometer inputs to the board. And finally, J1 and J2 are the inputs of motors to the board. In order to drive motors, one L298 is used that each IC provides the required startup current for the two motors. ATMEGA16 is used as the controller. The PCB of the electronic circuit is designed and implemented using Altium Designer.

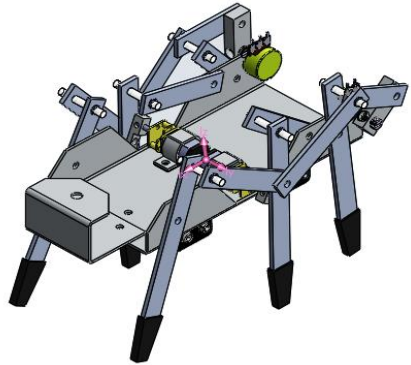

Figure 4 Center of mass of the robot

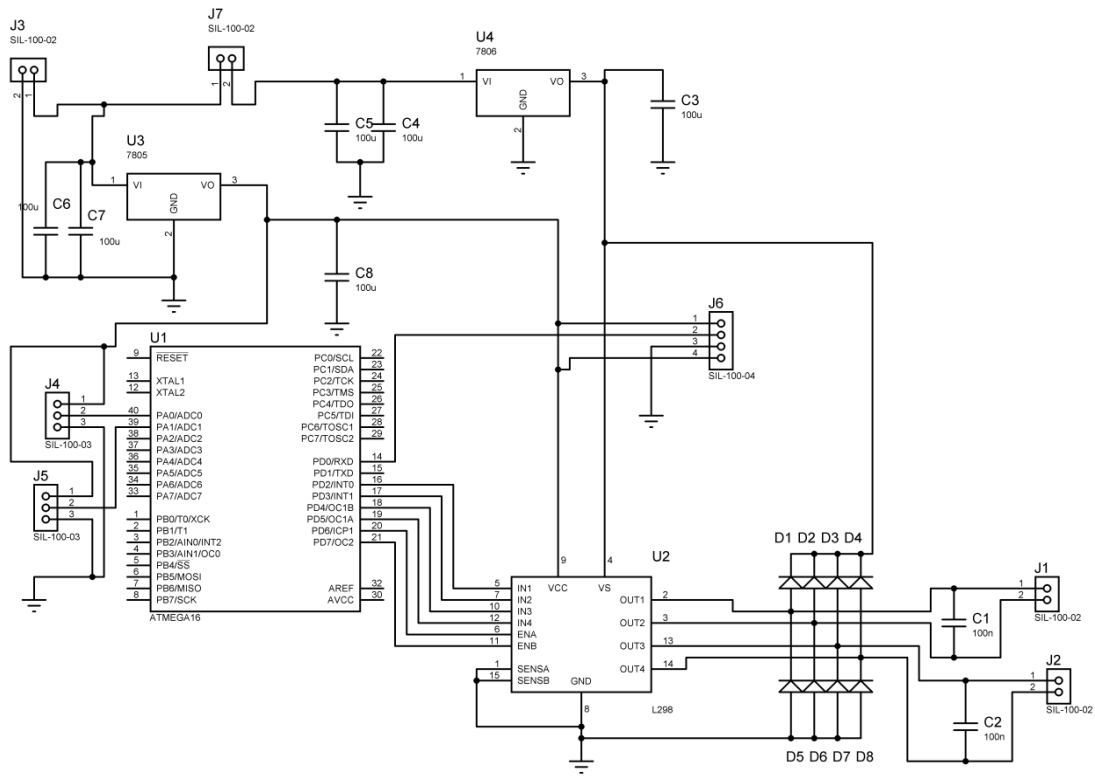

Figure 5 Electronic circuit designed in Proteus Professional

The microcontroller is programmed using Codevision. For wireless communication, two modules are used. The two modules are HM-T (Transmitter) and HM-R (Receiver). These modules are suitable due to small dimensions, low power consumption and appropriate range (up to 150 m). In order to connect the receiver module to the microcontroller, the DATA pin of the HM-R is directly connected to RXD (Receiver in USART protocol). To connect the transmitter to a supervisor computer a USB to serial converter is used. The transmitter interface is programmed using LabVIEW. One of the main issues of these modules is that the receiver is highly susceptible to noise and if every 75ms no data is sent to the receiver, the receiver will receive the noise and pass it on to the microcontroller. To overcome the mentioned issue, while not sending any data, the transmitter will send a neutral data to avoid the receiver from getting noise. Besides, in order to be sure that the sent data is actually the data that is intended to send and is not noise or data from other wireless devices, a password is assigned in the receiver's program. So if another data is sent from the surrounding modules simultaneously, the program will identify it and will not process it. On the other hand, encrypting the sent and received data will significantly reduce the sensitivity and susceptibility to noise, so that even if no neutral data is sent and the wrong data is given to microcontroller by the receiver, the program will identify that this data is noise and no action will be taken. Another faced issue was that the sent data from computer was altered by USB to serial convertor. So to solve this issue, first the altered data is diagnosed and second the altered data is considered in the microcontroller programming. Five modes of operation are defined for the robot. These modes are forward movement, backward movement, right turn, left turn and stop. These 5 modes of operation are controlled using LabVIEW. Figure 6 shows the interface of the program in LabVIEW.

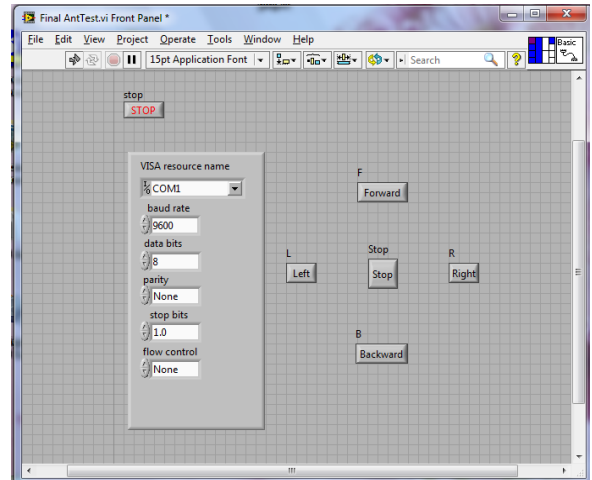

Figure 6 Modes of operation in LabVIEW

Figure 7 shows the transmitter module and Figure 8 demonstrates the overall system with the supervising computer and Figure 9 depicts the implemented six-legged ant robot.

The wireless algorithm is as follows:
Wireless Send Receive pseudo code:
Computer (Sender or Server) Algorithm:
While (start)
Send 1;
If (Command)
    Send Command
End if
End While

Ant (Receiver or Client) Algorithm:
While (1)
Receive Input
If (Input Is Command)
    Stop Previous Command
    Do New Command
End if
End While

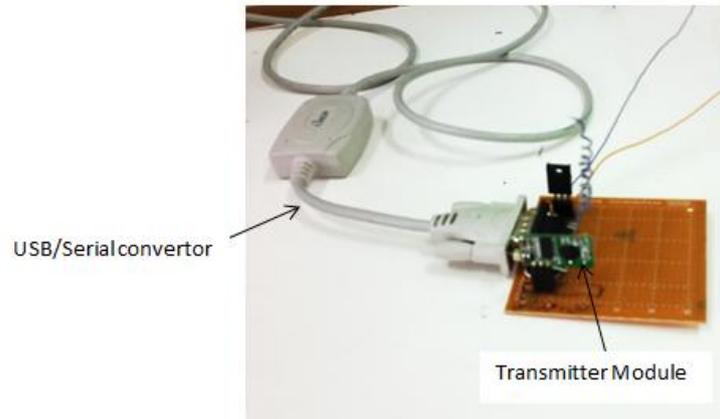
Figure 7 Transmitter module

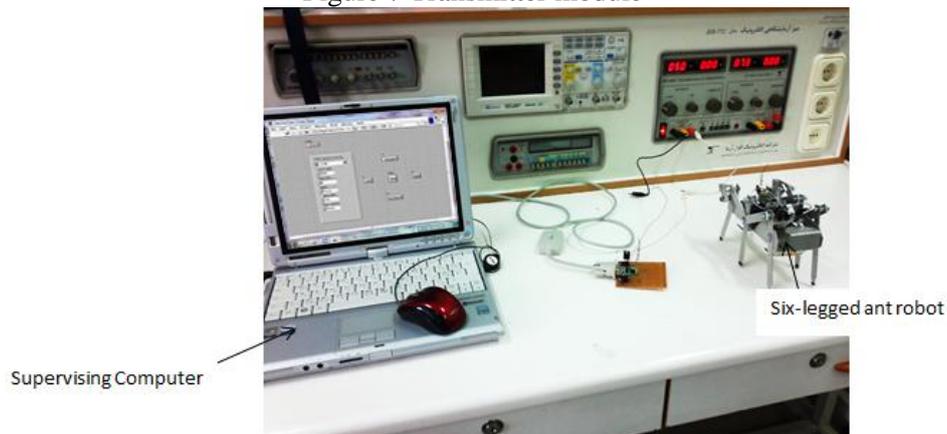
Figure 8 Overall system with supervising computer

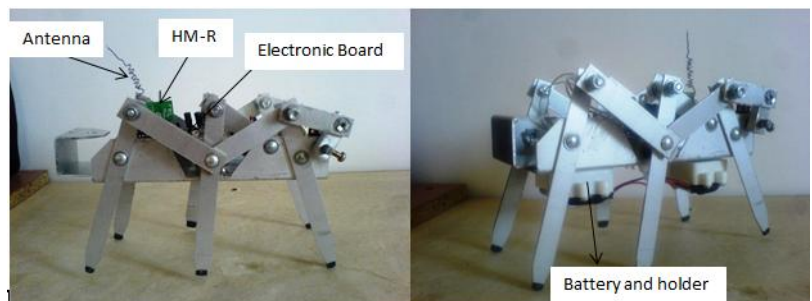
Figure 9 Six-legged AntBot I

**4. Discussion**

Living organisms have muscles that are more stronger than electric motors. The body organs are also stronger and lighter than synthetic materials. Ants are also very small creatures and their bodies are very delicate. Considering all these features put constraints on designing robots similar to living organisms. The closest artificial alloy that is both strong and light is aluminum. Ants are fascinating creatures since (1) they have six legs that this feature makes them stable creatures and so a six-legged robot will be a stable one; (2) they are strong and powerful creatures due to their body structure and they can carry heavy loads; (3) they are sociable and work together to carry out specific tasks and this goal can be achieved by making several ant robots; (4) they are creatures that can communicate with each other chemically, however this feature cannot be fully mimicked. But the most important thing is to make robots that can communicate and to achieve this goal, wireless communication is utilized.

A lot of research has been conducted on ant robots and more generally on six-legged robots, but the main advantages of this robot compared to others are: (1) being light-weight, (2) being low-cost, (3) ease of control, (4) having relatively high speed compared to fellow robots and (5) the ability to carry high loads with respect to its weight. All of these features together cannot be found in just one robot. Table1 shows a simple comparison between some six-legged robots with our AntBotI. From this table it can be seen that by considering the dimensions of the robot, it has high speed and load carrying capacity.

Table 1 Comparison of AntBot I with other six-legged robots.

| Robot | AntBot I | BILL-ANT-p | Lauron III | Robot II | Tarry II | TUM Walking Machine |
|---|---|---|---|---|---|---|
| Length(m) | 0.155 | .33 | 0.5 | 0.5 | 0.5 | 0.8 |
| Width(m) | 0.10 | .33 | 0.3 | 0.25 | 0.2 | 0.4 |
| Height(m) | 0.10 | .15 | 0.8 | 0.5 | 0.4 | 1.0 |
| Max. Speed(m/s) | 0.1 | 0.004 | 0.4 | 0.14 | 0.2 | 0.3 |
| Weight(kg) | 0.230 | 2.3 | 18 | -- | 2.9 | 23 |
| Load(kg) | .075 | 8.6 | 10 | -- | 2.9 | 5 |
| Legs | 6 | 6 | 6 | 6 | 6 | 6 |
| Power Consumption(W) | 0.9 | 20 | 90 | -- | 30 | 500 |

## 5. Conclusion

In this paper, design, simulation and implementation of a new six-legged ant robot was proposed. The kinematic structure of the robot was parallel and only two motors were used to actuate the six legs. The mechanical structure was designed using Solidworks. The robot can walk and turn and because of using wireless modules it is semiautonomous. The transmitter program was implemented using LabVIEW on the supervisor computer and five modes of operation were defined for the robot. The robot is light weight and can mimic the ant.

## 6. Future work

There are several possible extensions that can be made to this robot. First a vision system can be implemented and a vision-based control strategy can be used. The vision system can be put on the head of the robot. Second, obstacle-avoidance sensors such as infrared can be used to avoid obstacles and position and velocity sensors such as encoders can be used instead of potentiometers. Additionally by considering the fact that the kinematic structure of the robot is parallel [19], various control strategies can be implemented to help the robot walk even on rough terrain. However, it should be noted that the aim of AntBot I was to make a six-legged robot and the control strategy will be implemented in the future. Since the robot is equipped with a wireless system, in the future the robot will communicate with fellow ants and by their contribution it will be able to carry out a task. This can be done by either using a central server and ants as clients or all ants can communicate with each other without the intervention of a central server. The simplicity of this robot makes the robot both cost-effective and suitable for social applications.